\documentclass[11pt]{article}

\usepackage[margin=1in]{geometry}
\usepackage[T1]{fontenc}
\usepackage[utf8]{inputenc}
\usepackage{lmodern}
\usepackage{microtype}
\usepackage{amsmath,amssymb}
\usepackage{booktabs}
\usepackage{graphicx}
\usepackage{caption}
\usepackage{subcaption}
\usepackage{xcolor}
\usepackage{placeins}
\usepackage{hyperref}
\usepackage{tikz}
\usepackage{cite}
\usetikzlibrary{arrows.meta,positioning,fit}

\newcommand{\prauc}{PR--AUC}
\newcommand{\rocauc}{ROC--AUC}

\title{\textbf{Quantum-Inspired Contextual Learning for Sparse-Ring Fraud Detection in Dynamic Transaction Graphs}}
\usepackage{authblk}
\author[1]{Behnam Tonekaboni}
\author[2]{Hiroshi Yamauchi}
\affil[1]{Infleqtion Australia, Melbourne}
\affil[2]{SoftBank Corp., Japan, Tokyo}

\begin{document}

\date{}
\maketitle

\begin{abstract}
We present an exploratory benchmark and quantum-inspired modeling prototype for fraud screening in dynamic financial transaction graphs. Coordinated fraud may not be visible from individual transactions alone, but may emerge as a multi-period relational pattern. We focus on sparse-ring fraud, a stylized pattern in which a completed directed cycle is distributed across several days, requiring models to integrate evidence across both time and graph structure. We study this problem using a synthetic transaction simulator with completed sparse-ring injections and broken-ring decoys. Daily directed transaction graphs are aggregated into rolling windows and represented using raw graph features, persistent-homology summaries, or hybrid feature vectors that combine both. We compare a gated recurrent unit (GRU) baseline with quantum-inspired Contextual Machine Learning (CML) as sequence-level classifiers. Because the benchmark uses synthetic data, a modest sample size, and sequence-level labels, the results are exploratory. Within this scope, topology-only summaries are too compressed to solve the supervised ring-completion task by themselves, largely because they remove account-pair identity and edge direction. The strongest results come from hybrid representations that combine identity-preserving graph features with topological summaries. These findings suggest that topology is most useful as a contextual layer over dynamic graph features, and that CML is a promising candidate model for fraud patterns whose evidence is distributed across temporal and relational context.
\end{abstract}

\noindent\textbf{Keywords:} Fraud detection; dynamic transaction graphs; sparse-ring fraud; quantum-inspired machine learning; Contextual Machine Learning; topological data analysis; persistent homology; sequence classification.

\section{Introduction}

Financial fraud is often both relational and temporal. A single transaction may appear ordinary when viewed in isolation, while a sequence of transactions among multiple accounts may reveal coordinated movement of funds, layering behavior, circular transfers, or other suspicious structures. This motivates a dynamical graph-based view of fraud detection, in which accounts are represented as vertices, transactions as directed weighted edges, and suspicious behavior as a pattern in a dynamic transaction network~\cite{pandit2007netprobe,cheng2024gnnfraudreview}.

In this paper we use a controlled benchmark to study fraud detection in dynamic transaction graphs. We first asks whether a completed sparse ring---a small directed cycle whose edges are distributed across several days---can be detected as a temporal graph pattern. Then we examines two related comparative questions. First, we ask whether quantum-inspired Contextual Machine Learning (CML)~\cite{infleqtionCML} can detect fraud patterns whose evidence is distributed across time and relational structure as effectively as, or better than, a conventional recurrent sequence model such as a gated recurrent unit (GRU). Second, we ask whether topological summaries of transaction graphs provide useful additional context for this detection task. Although the sparse-ring pattern is intentionally simplified, it captures an important detection challenge. Each edge may resemble an ordinary transaction when observed within a single day, while the fraud signal becomes apparent only after evidence is integrated across a rolling temporal window.

This setting motivates the use of quantum-inspired CML~\cite{infleqtionCML} together with topology-based graph features. We treat CML as a sequence-level classifier for patterns whose significance depends on temporal and relational context, and we use topological summaries to test whether graph shape adds useful information beyond raw transaction features. The goal is therefore to compare CML with a GRU baseline and to evaluate whether topological augmentation improves detection of temporally distributed sparse rings.

The quantum component of this study is methodological rather than hardware-based. In the present experiments, we use CML as a classical, quantum-inspired model motivated by quantum contextuality and quantum correlations; the results should not be interpreted as evidence of hardware-level quantum advantage. Nevertheless our study intended to be relevant to future quantum-native extensions, because sparse-ring detection involves temporal context, graph cycles, and higher-order relational structure. These features connect naturally to future work on quantum contextual learning~\cite{gao2022enhancing,PRXQuantum.4.020338,anschuetz2026arbitrary}, quantum topological data analysis (QTDA)\cite{lloyd2016quantum,mcardle2026streamlined}, and quantum–topological signal processing (QTSP)\cite{leditto2025tsp,leditto2025quantum}.

\subsection{Research scope and contributions}

In this paper, we present an exploratory simulation study of sparse-ring fraud detection in synthetic dynamic transaction graphs. The simulator allows the timing and structure of completed sparse rings and broken-ring decoys to be controlled precisely, making it possible to compare feature representations and model classes under matched conditions. We use this benchmark to evaluate raw graph features, topology-only summaries, and hybrid representations, and to compare a GRU baseline with quantum-inspired CML. The study is an early-stage methodological prototype: a controlled benchmark for testing sparse-ring detection, topological augmentation, and GRU--CML model comparison under synthetic conditions. Its current scope is defined by a modest benchmark size, limited uncertainty analysis across repeated runs, sequence-level labels, and simulated data that have not yet been calibrated against real transaction patterns.

Our first contribution is a controlled dynamic-graph simulator for studying fraud patterns that unfold over time. The simulator creates account-level transaction activity with different types of account behavior and day-to-day variation. It then injects sparse-ring patterns, where each individual transfer may look normal if viewed on a single day. The benchmark also includes broken-ring decoys, which makes the task more meaningful: a good model should not only detect unusual activity, but also recognize whether the activity forms a complete coordinated structure over time.

The second contribution is a graph-to-topology feature pipeline designed for rolling transaction windows. Daily directed transaction graphs are aggregated over time, converted into distance-like representations, and summarized using persistent homology in dimensions zero and one~\cite{edelsbrunner2002,zomorodian2005,ghrist2008,carlsson2009}. This pipeline makes it possible to ask whether topological descriptors of connectivity and loop-like structure provide information that is not captured by conventional transaction aggregates alone.

The third contribution is an exploratory comparison between a GRU baseline and Infleqtion CML, a quantum-inspired contextual model, across raw, topology-only, and hybrid feature representations. This comparison tests whether CML is competitive with a standard recurrent sequence model when the fraud label depends on multi-period graph structure rather than on isolated transaction features.

\section{Benchmark design and research propositions}

\subsection{Problem formulation}

As described in the introduction, and following graph-based approaches to fraud detection~\cite{pandit2007netprobe,cheng2024gnnfraudreview}, we model daily transaction activity as a sequence of directed weighted graphs. For each day, we represent accounts as vertices and transactions as directed edges, and we use edge weights to record quantities such as transaction amount. We aggregate these daily graphs into rolling histories and train each model to assign a sequence-level score to each history. The label indicates whether the history contains a completed sparse-ring pattern. We keep the task at the sequence level so that we can first test whether temporal graph structure and topological summaries carry useful signal before moving to account-level or transaction-level alerts.

For each day $t \in \{1,\ldots,T\}$, let
\begin{equation}
    G_t = (V, E_t, w_t)
\end{equation}
denote the directed weighted transaction graph observed on that day. The vertex set $V$ represents accounts, $E_t \subseteq V \times V$ represents directed transactions, and $w_t(i,j)$ records an edge attribute such as total amount or transaction count from account $i$ to account $j$. The dynamic transaction record over a horizon of $T$ days is
\begin{equation}
    \mathcal{G}_{1:T} = (G_1, G_2, \ldots, G_T).
\end{equation}

The implemented benchmark is a sequence-level detection problem. Let $s \in \{1,\ldots,N\}$ index a simulated transaction-history sequence, where $N$ is the number of simulated graph histories in the dataset. For sequence $s$, the full dynamic graph record is
\begin{equation}
    \mathcal{G}^{(s)}_{1:T}
    =
    \left(G^{(s)}_1,\ldots,G^{(s)}_T\right).
\end{equation}

Rather than classifying a single transaction or a single daily graph, the model receives a time-ordered representation of the graph history. For each window $k$, let
\begin{equation}
    \mathcal{G}^{(s)}_{a_k:b_k}
    =
    \left(G^{(s)}_{a_k},G^{(s)}_{a_k+1},\ldots,G^{(s)}_{b_k}\right)
\end{equation}
denote the subsequence of daily transaction graphs from start day $a_k$ to end day $b_k$. In the simplest case, all windows have fixed length $L$ days and are advanced by a step size $\delta$. Then
\begin{subequations}
\begin{align}
    a_k &= 1 + (k-1)\delta, \\
    b_k &= a_k + L - 1,
    \qquad k = 1,\ldots,K.
\end{align}
\end{subequations}
The number of windows is
\begin{equation}
    K = \left\lfloor \frac{T-L}{\delta} \right\rfloor + 1.
\end{equation}
Daily feature sequences are recovered as the special case $L=1$ and
$\delta=1$, for which $K=T$.

Each window is mapped to a feature vector
\begin{equation}
    x^{(s)}_k
    =
    \Phi\!\left(\mathcal{G}^{(s)}_{a_k:b_k}\right),
    \qquad
    k = 1,\ldots,K,
\end{equation}
where $\Phi(\cdot)$ is the feature extraction map. Depending on the experiment, $\Phi$ may compute raw graph features, topological summaries, or a hybrid representation. The resulting model input is the ordered feature sequence
\begin{equation}
    X^{(s)} = \left(x^{(s)}_1,\ldots,x^{(s)}_K\right).
\end{equation}

The supervised target used in the current experiments is a binary sequence-level label. For each simulated sequence $s$, we define
\begin{equation}
    y^{(s)}
    =
    \mathbf{1}\left\{
    \mathcal{G}^{(s)}_{1:T}
    \text{ contains a completed sparse directed ring}
    \right\}.
\end{equation}
Equivalently, $y^{(s)}=1$ means that the sequence contains all directed edges needed to complete the sparse ring pattern within the observation horizon, while $y^{(s)}=0$ means that no completed sparse directed ring is present.

Sequences with $y^{(s)}=0$ can arise in two ways. In the clean case, no sparse-ring-like event is injected. In the decoy case, ring-like directed transfers are injected, but they do not close into a completed directed ring. These broken-ring decoys are not treated as positive fraud labels in the current supervised task. Instead, they serve as hard comparison cases for testing whether a model can distinguish a completed loop from incomplete ring-like activity.

Given the input sequence $X^{(s)}$, the trained model returns a score
\begin{equation}
    \hat{p}^{(s)} = f_\theta\!\left(X^{(s)}\right) \in [0,1].
\end{equation}
A larger value of $\hat{p}^{(s)}$ means that the model considers sequence $s$ more likely to contain a completed sparse directed ring. The score is therefore evaluated against the sequence-level label $y^{(s)}$.

For sequences containing broken-ring decoys, a high score is treated as a false alarm, because these sequences do not contain a completed ring. This allows us to measure whether the model is detecting ring completion rather than simply reacting to incomplete ring-like activity.

We deliberately keep this formulation at the sequence level. In an operational extension, we would introduce finer-grained account-level or edge-level labels, for example
\begin{equation}
    y_{i,t}
    \quad \text{or} \quad
    y_{(i,j),t},
\end{equation}
to link each alert to specific accounts, transactions, or local subgraphs. In the present experiments, we do not train on these entity-level or edge-level targets. Instead, we use the sequence-level task as a controlled first step to test whether temporal graph structure and topological summaries contain useful signal for detecting completed coordinated fraud patterns.

\subsection{Research propositions}

We organize the benchmark around three Research Propositions (RPs). Each RP links the sequence-level label $y^{(s)}$ to the model score $\hat{p}^{(s)}$ and specifies how we evaluate sparse-ring detection, topological augmentation, and GRU--CML model comparison. We split the simulated sequences into training, validation, and held-out test sets. We fit model parameters on the training set, use the validation set to tune model settings and select thresholds, and reserve $\mathcal{D}_{\mathrm{test}}$ for final evaluation.

For each model choice $m$ and feature representation $r$, we assign each test sequence a score 
$\hat{p}^{(s)}_{m,r}$. The model index $m$ denotes the classifier used to map a feature sequence to a score, such as GRU or CML. The representation index $r$ denotes the input representation used to construct $X_r^{(s)}$, such as raw graph features, topological features, or hybrid features. We evaluate how well these scores rank completed-ring sequences above sequences without completed rings using two standard ranking metrics, ROC-AUC and PR-AUC~\cite{fawcett2006roc,saito2015precision}. ROC-AUC measures whether completed-ring sequences tend to receive higher scores than non-completed-ring sequences across decision thresholds. PR-AUC summarizes the precision--recall tradeoff and is especially useful when positive cases are rare or when false positives affect the usefulness of alerts. We denote these metrics by $A_{\mathrm{ROC}}(m,r)$ and $A_{\mathrm{PR}}(m,r)$, respectively.

We also measure how often each model assigns a high score to broken-ring decoys. Let $\mathcal{D}_{\mathrm{decoy}}$ be the subset of test sequences that contain a broken sparse-ring decoy but no completed ring. For a threshold $\tau_{m,r}$ selected on the validation set, we define the empirical decoy false-alarm rate as
\begin{equation}
    \mathrm{FAR}_{\mathrm{decoy}}(m,r;\tau_{m,r})
    =
    \frac{1}{|\mathcal{D}_{\mathrm{decoy}}|}
    \sum_{s \in \mathcal{D}_{\mathrm{decoy}}}
    \mathbf{1}\!\left\{
        \hat{p}^{(s)}_{m,r} \geq \tau_{m,r}
    \right\}.
\end{equation}
A lower value means that the model is less likely to mistake an incomplete ring-like pattern for a completed fraud ring.

\paragraph{RP1: Can sparse fraud be detected as a temporal graph pattern?}

The first question is whether the model can detect a completed sparse ring when the ring edges occur on different days. For a completed ring in sequence $s$,
let
\begin{equation}
    d^{(s)}_1 < d^{(s)}_2 < \cdots < d^{(s)}_q
\end{equation}
denote the days on which the ring edges appear. The ring is only completed when the final edge appears. Before that point, the sequence may contain only partial ring-like activity. This makes the task different from detecting a single unusual transaction or a single abnormal daily graph.

We evaluate RP1 by asking whether completed-ring sequences, those with $y^{(s)}=1$, receive higher model scores than sequences without completed rings. This is summarized by $A_{\mathrm{ROC}}$ and $A_{\mathrm{PR}}$. We also measure recall at a threshold $\tau$ chosen on the validation set:
\begin{equation}
    \mathrm{Recall}_{\mathrm{ring}}(\tau)
    =
    \Pr\!\left(
        \hat{p}^{(s)} \geq \tau
        \mid
        y^{(s)}=1
    \right).
\end{equation}
Here, recall means the fraction of true completed-ring sequences that the model scores above the threshold $\tau$. 

To test the temporal-memory aspect more directly, we group completed-ring sequences by how far apart the ring edges are. Let
\begin{equation}
    g^{(s)}
    =
    \max_j \left(d^{(s)}_{j+1} - d^{(s)}_j\right)
\end{equation}
be the largest gap between consecutive ring edges in sequence $s$. We then measure recall conditional on this gap:
\begin{equation}
    \mathrm{Recall}_{\mathrm{ring}}(g;\tau)
    =
    \Pr\!\left(
        \hat{p}^{(s)} \geq \tau
        \mid
        y^{(s)}=1,\; g^{(s)}=g
    \right).
\end{equation}

RP1 is supported if the model ranks completed-ring sequences above non-completed-ring sequences and maintains meaningful recall when the ring edges are separated across time. A sharp drop in recall for larger gaps would suggest that the model is relying mainly on short-term local cues rather than learning the full temporal ring pattern.

\paragraph{RP2: Do topological features add useful structural context?}

The second question is whether topological features provide information that is not already captured by the raw transaction features. Raw graph features keep more direct information about accounts, directed edges, transaction counts, and amounts. Topological features instead summarize the shape of a rolling graph window, such as connected components, loop-like structure, and connectivity across distance scales. These summaries may help when fraud is expressed as a coordinated graph pattern, but they may also lose useful account-level detail.

We test RP2 by comparing three representations of the same graph-history sequence:
\begin{equation}
    r \in \{\mathrm{raw},\mathrm{topology},\mathrm{hybrid}\}.
\end{equation}
The raw representation uses transaction-graph features only. The topology representation uses topological summaries only. The hybrid representation combines both.

The main comparison is whether adding topological features improves ranking performance over raw graph features alone. For model $m$, we define the hybrid PR-AUC lift as
\begin{equation}
    \Delta_{\mathrm{hybrid}}(m)
    =
    A_{\mathrm{PR}}(m,\mathrm{hybrid})
    -
    A_{\mathrm{PR}}(m,\mathrm{raw}).
\end{equation}
Here, $\Delta_{\mathrm{hybrid}}(m)>0$ means that the hybrid representation gives higher PR-AUC than the raw representation for model $m$.

RP2 is supported if the hybrid representation improves performance over raw features, especially in decoy settings. This would suggest that topological summaries add useful structural context for distinguishing completed rings from incomplete ring-like activity. If the topology-only representation performs poorly but the hybrid representation improves over raw features, that would suggest that topology is more useful as complementary context than as a standalone replacement for account- and edge-level information.

\paragraph{RP3: Can quantum-inspired CML exploit temporal and topological context?}

The third question is whether CML is competitive with a standard recurrent baseline on the same sequence-level detection task. The GRU baseline is included because recurrent neural networks are a common choice for ordered sequences. CML is included because the task depends on context: the meaning of one edge may depend on other edges that appeared earlier or later in the graph history.

We compare CML and GRU using the same train/validation/test splits, the same sequence-level labels, and the same feature representations. This keeps the comparison focused on the modeling approach rather than differences in data or features.

For each representation $r$, we measure the PR-AUC difference between CML and GRU:
\begin{equation}
    \Delta_{\mathrm{CML}}(r)
    =
    A_{\mathrm{PR}}(\mathrm{CML},r)
    -
    A_{\mathrm{PR}}(\mathrm{GRU},r).
\end{equation}
Here, $\Delta_{\mathrm{CML}}(r)>0$ means that CML gives higher PR-AUC than the GRU for representation $r$. A value near zero means that the two models perform similarly, while a negative value means that the GRU performs better.

We also compare how the two models behave on broken-ring decoys. Let $\mathrm{FAR}_{\mathrm{decoy}}(m,r;\tau)$ denote the false-alarm rate on decoy  sequences for model $m$ and representation $r$ at threshold $\tau$. We define
\begin{equation}
    \Delta_{\mathrm{decoy}}(r)
    =
    \mathrm{FAR}_{\mathrm{decoy}}(\mathrm{CML},r;\tau)
    -
    \mathrm{FAR}_{\mathrm{decoy}}(\mathrm{GRU},r;\tau).
\end{equation}
Here, $\Delta_{\mathrm{decoy}}(r)<0$ means that CML produces fewer false alarms on broken-ring decoys than the GRU. Conversely, a positive value means that CML is more likely than the GRU to mistake decoys for completed rings.

RP3 is supported if CML achieves competitive or higher PR-AUC while maintaining similar or lower false-alarm rates on broken-ring decoys. The goal is to test whether CML is a promising contextual classifier for fraud patterns where temporal memory and graph structure are central.

\section{Transaction Simulation}

We generate controlled sequences of daily directed transaction graphs to create a transparent benchmark for sparse-ring fraud detection. We first simulate normal background transaction activity and then inject coordinated events with known timing, participants, and labels. We include both completed multi-day rings and incomplete ring-like decoys. This design lets us test whether a model can integrate evidence across days and distinguish true ring completion from normal background activity and decoy structures.

For each simulated sequence, we first create a fixed set of account nodes $V$ and initialize $T$ empty daily graph snapshots,
\begin{equation}
    G_t^{(0)} = (V,\emptyset),
    \qquad t=1,\ldots,T.
\end{equation}
We then populate these daily graphs with simulated transactions. On day $t$, whenever account $i$ sends at least one transaction to account $j$, we add the directed edge $i \rightarrow j$ to $G_t$. If the same ordered pair transacts multiple times on the same day, we aggregate those transactions into a single edge. For each edge, we store the transaction count and the total transaction amount as edge attributes.

\subsection{Normal transaction dynamics}
We simulate normal transaction activity before injecting any sparse-ring events. The background process creates dynamic graphs with role structure, temporal persistence, periodic bursts, heterogeneous transaction amounts, and random noise.

We assign each account one of four roles: salary-like source, merchant-like receiver, higher-activity hub, or ordinary retail account. These roles determine the baseline propensity of transactions between ordered account pairs. For example, retail accounts pay merchant accounts with higher probability, merchant accounts interact with hubs more often, and hubs have broader outgoing activity. We also allow low-probability interactions between other role pairs so that the graph contains incidental background activity.

For each day $t$, we sample the number of background transactions from a Poisson distribution,
\begin{equation}
    N_t \sim \mathrm{Poisson}(\lambda_t).
\end{equation}
We set the daily rate $\lambda_t$ using a baseline transaction volume, a weekday/weekend multiplier, and day-level noise. This produces denser and sparser daily graphs across the simulated horizon.

We generate each background transaction in three steps. First, we select a sender account. Second, we sample a receiver account using the role-dependent propensity matrix. Third, we draw the transaction amount from a log-normal distribution, which produces positive and right-skewed transaction sizes. When sampling the receiver, we also include two temporal effects: we increase the probability of an ordered pair if that pair transacted on the previous day, and we increase salary-to-retail flows on periodic payday days.

After generating all transactions for day $t$, we aggregate repeated transactions between the same ordered account pair. Each directed edge therefore stores two attributes: the transaction count and the total amount exchanged on that day.

Table~\ref{tab:simulation-components} summarizes the main mechanisms that we use to generate normal background transaction activity before injecting sparse-ring events. Together, these components produce structured but stochastic transaction histories with role-dependent behavior, daily volume variation, temporal persistence, periodic bursts, heterogeneous edge weights, and background noise. After generating this background activity, we inject completed sparse rings or broken-ring decoys and assign the sequence label according to the injected event type. This design ties the supervised label to the coordinated ring mechanism while preserving realistic variation in the surrounding transaction history.

\begin{table}[tbp]
\centering
\caption{Main components of the normal transaction simulator.}
\label{tab:simulation-components}
\begin{tabular}{p{0.28\linewidth} p{0.62\linewidth}}
\hline
\textbf{Component} & \textbf{Role in the simulation} \\
\hline
Account roles & Create structured sender--receiver behavior across salary, merchant, hub, and retail accounts. \\
Daily volume variation & Produces denser and sparser days through calendar effects and random day-level noise. \\
Temporal persistence & Makes recently active ordered pairs more likely to transact again. \\
Payday effect & Adds periodic bursts in salary-to-retail flows. \\
Log-normal amounts & Produces positive, right-skewed transaction amounts. \\
Edge aggregation & Converts repeated same-day transactions into edge counts and total amounts. \\
\hline
\end{tabular}
\end{table}

\subsection{Sparse-ring fraud mechanism}

Our benchmark focuses on sparse directed rings as the stylized fraud mechanism. In this section, we define a completed sparse ring and explain how we inject it into the simulated transaction history. For three ring participants $(v_1,v_2,v_3)$, a completed ring contains the directed edges
\begin{equation}
    v_1 \rightarrow v_2, \qquad
    v_2 \rightarrow v_3, \qquad
    v_3 \rightarrow v_1.
    \label{eq:completed-sparse-ring}
\end{equation}

We make the ring sparse by placing these edges on different days rather than on the same daily graph. For example, one edge may occur on day 3, another on day 7, and the closing edge on day 10. These day numbers are illustrative; in the simulator, we choose the edge days according to the benchmark setting.

A completed sparse ring is therefore a temporal graph pattern. A single daily graph may contain only one edge of the ring, or no ring edge at all. The pattern becomes visible only when a model integrates evidence across the observation horizon. This is the central reason the benchmark tests temporal memory rather than only single-day anomaly detection.

\begin{figure}[htbp]
    \centering
    \includegraphics[width=0.8\linewidth]{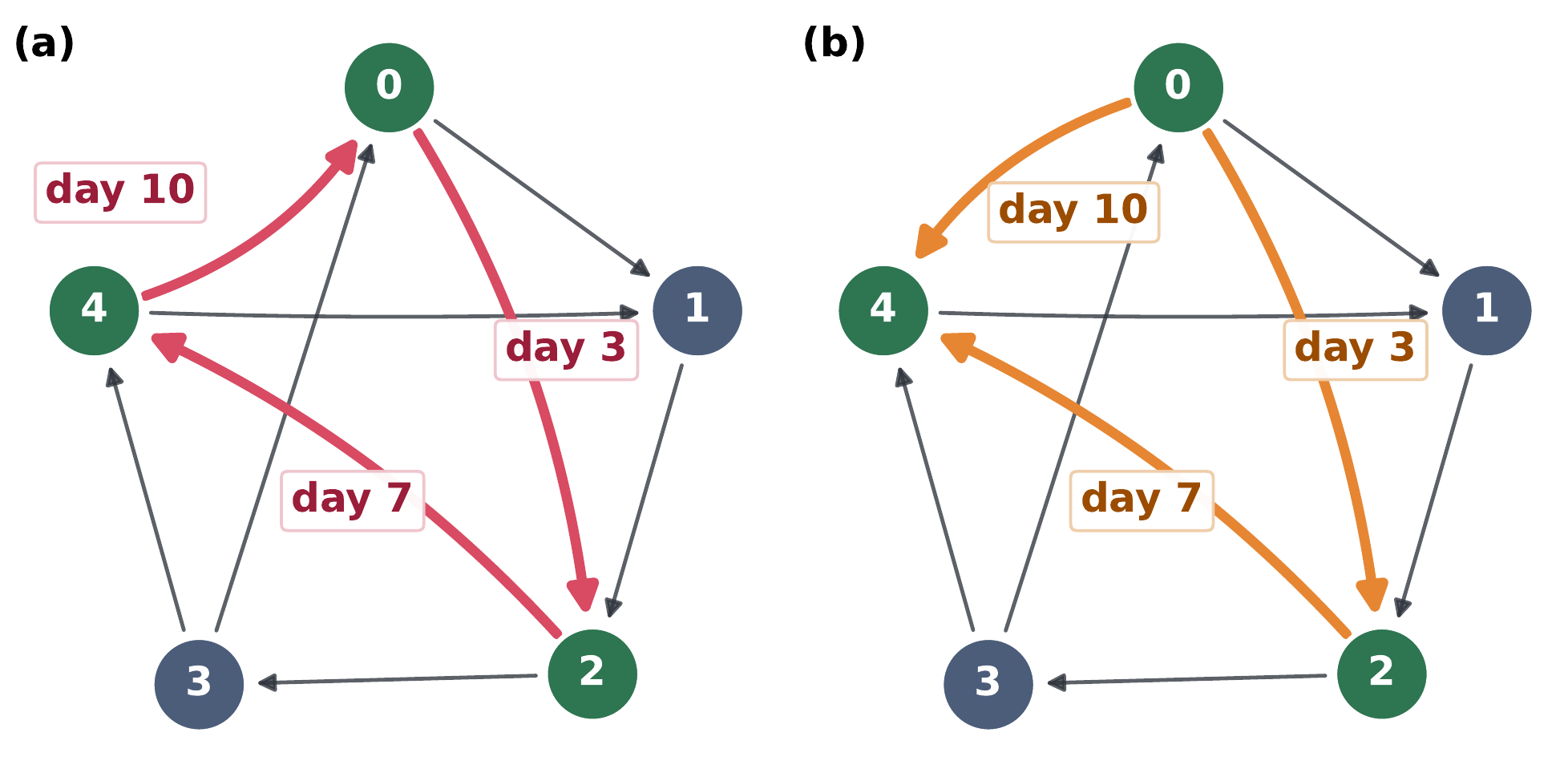}
    \caption{Sparse-ring and broken-ring transaction patterns. 
(a) A completed sparse directed ring, where the ring edges occur on different days but close into a directed cycle over the full observation horizon. 
(b) A broken-ring decoy, where scheduled ring-like transfers are present but the final edge does not close the directed cycle. Decoy sequences therefore contain suspicious sparse activity but are labeled $y=0$ because no completed ring is present.}
    \label{fig:sparse-ring}
\end{figure}

Figure~\ref{fig:sparse-ring}(a) illustrates the distinction between a completed sparse ring and a broken-ring decoy. In both cases, the highlighted transfers are spread across multiple days, but only the completed sparse ring closes the directed cycle.

To inject a completed sparse ring, we select the ring participants, assign one day to each ring edge, and add the corresponding scheduled transactions to the appropriate daily graphs. We record these injected transactions in the same way as background transactions: they contribute to the observed edge count and total transaction amount for that ordered account pair on that day. We assign the sequence label $y=1$ because the observation horizon contains a completed directed ring.

\subsection{Broken-ring decoys}

We include broken-ring decoys as hard negative cases for the sparse-ring benchmark. These decoys use the same scheduling mechanism as completed sparse rings: we inject coordinated transfers across multiple days, with similar timing and transaction structure. The key difference is that the injected transfers form an incomplete ring-like pattern rather than the completed directed cycle in Equation~\eqref{eq:completed-sparse-ring}.

To construct a broken-ring decoy, we keep the same three participants but replace the closing transfer with a non-closing directed edge. For example, instead of closing the cycle with $v_3 \rightarrow v_1$, we inject
\begin{equation}
    v_1 \rightarrow v_2,\qquad
    v_2 \rightarrow v_3,\qquad
    v_1 \rightarrow v_3.
    \label{eq:broken-ring-decoy}
\end{equation}
Figure~\ref{fig:sparse-ring}(b) illustrates this construction. The highlighted transfers are still spread across multiple days, but they do not return to the starting account and therefore do not complete the directed cycle.

We generate decoy sequences by adding these scheduled non-closing transfers on top of the same normal background activity used for other sequences. We record the injected decoy transfers in the daily graphs in the same way as background transactions, so they contribute to the observed edge counts and total transaction amounts. Since the supervised target marks completed directed rings, we assign each broken-ring decoy sequence the label $y=0$.

The decoys make the benchmark focus on ring completion. Because both positive sequences and decoy sequences contain scheduled sparse transfers among related accounts, a model must use the temporal and directed graph structure to distinguish completed rings from incomplete ring-like patterns.

\subsection{Dataset variants}

We construct three dataset variants by keeping the positive class fixed and changing the negative class. In every variant, sequences with completed sparse rings receive the label $y=1$. Throughout these variants, $y=0$ means that the sequence does not contain a completed sparse directed ring; it does not necessarily mean that the sequence contains only ordinary background activity. We vary the $y=0$ sequences to control how strongly the benchmark tests ring completion rather than general abnormal activity.

\begin{itemize}
    \item \textbf{Easy-clean:} We use normal background sequences as the $y=0$ class. This variant serves as a sanity check: it tests whether the model can detect the presence of an injected completed sparse ring against ordinary background activity.

    \item \textbf{Hard-decoy:} We use broken-ring decoy sequences as the $y=0$ class. These sequences contain scheduled ring-like transfers, but the directed cycle does not close. This variant directly tests whether the model detects ring completion rather than merely responding to sparse injected activity.

    \item \textbf{Mixed:} We use a mixture of normal background sequences and broken-ring decoy sequences as the $y=0$ class. This variant combines clean negatives with harder decoy negatives and provides an intermediate benchmark regime.
\end{itemize}

\section{Graph-to-topology feature pipeline}

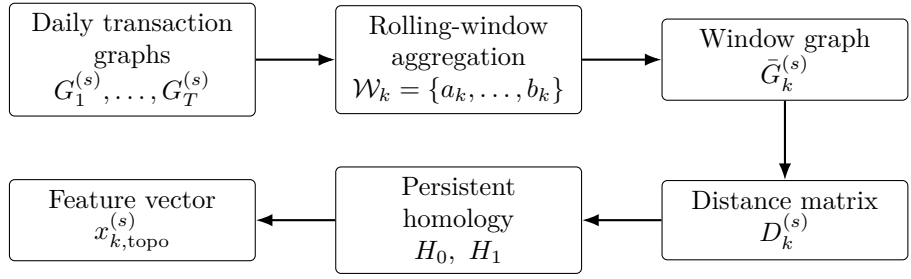
\begin{figure}[tbp]
    \centering
    \begin{tikzpicture}[
    node distance=1.05cm and 1.05cm,
    stage/.style={
        draw,
        rounded corners=2pt,
        align=center,
        minimum width=3.25cm,
        minimum height=1.05cm,
        font=\small
    },
    arrow/.style={
        -{Latex[length=2.2mm]},
        thick
    }
]

\node[stage] (daily) {
    Daily transaction\\
    graphs\\
    $G^{(s)}_1,\ldots,G^{(s)}_T$
};

\node[stage, right=of daily] (window) {
    Rolling-window\\
    aggregation\\
    $\mathcal{W}_k=\{a_k,\ldots,b_k\}$
};

\node[stage, right=of window] (aggregate) {
    Window graph\\
    $\bar{G}^{(s)}_k$
};

\node[stage, below=of aggregate] (metric) {
    Distance matrix\\
    $D^{(s)}_k$
};

\node[stage, left=of metric] (ph) {
    Persistent\\
    homology\\
    $H_0,\ H_1$
};

\node[stage, left=of ph] (features) {
    Feature vector\\
    $x^{(s)}_{k,\mathrm{topo}}$
};

\draw[arrow] (daily) -- (window);
\draw[arrow] (window) -- (aggregate);
\draw[arrow] (aggregate) -- (metric);
\draw[arrow] (metric) -- (ph);
\draw[arrow] (ph) -- (features);

\end{tikzpicture}
    \caption{Graph-to-topology feature pipeline. We aggregate daily directed transaction graphs into rolling windows, convert each window graph into a distance matrix, compute persistent homology, and summarize the resulting $H_0$ and $H_1$ persistence diagrams as fixed-length topology feature vectors.}
    \label{fig:pipeline}
\end{figure}

We use topological data analysis to convert rolling transaction graphs into fixed-length structural features. Persistent homology tracks topological features across filtration scales, such as connected components and loop-like structures~\cite{edelsbrunner2002,zomorodian2005,ghrist2008,carlsson2009}. For each rolling window, we aggregate the daily directed transaction graphs, convert the resulting window graph into a distance matrix, compute persistent homology, and summarize the resulting persistence diagrams as feature vectors. Figure~\ref{fig:pipeline} gives an overview of this graph-to-topology feature pipeline.

\subsection{Rolling-window aggregation}
For each simulated sequence $s$, we start with the dynamic graph record $\mathcal{G}^{(s)}_{1:T}=(G^{(s)}_1,\ldots,G^{(s)}_T)$, where each $G^{(s)}_t$ is the directed transaction graph for day $t$. To construct topology features, we aggregate the daily graphs over rolling windows of length $L$. We denote window $k$ by
\begin{equation}
    \mathcal{W}_k = \{a_k,\ldots,b_k\},
    \qquad
    b_k-a_k+1=L,
\end{equation}
for $k=1,\ldots,K$. Each window produces one aggregated graph and therefore one topology feature vector.

For each rolling window $\mathcal{W}_k$, we build one aggregated window graph
$\bar{G}^{(s)}_k$. We construct this graph by summing transaction activity
between each ordered account pair across all days in the window. Let
$C^{(s)}_{t,ij}$ denote the transaction count from account $i$ to account $j$
on day $t$, and let $A^{(s)}_{t,ij}$ denote the corresponding total amount.
We define the window-level count and amount as
\begin{equation}
    C^{(s)}_{k,ij}
    =
    \sum_{t \in \mathcal{W}_k} C^{(s)}_{t,ij},
    \qquad
    A^{(s)}_{k,ij}
    =
    \sum_{t \in \mathcal{W}_k} A^{(s)}_{t,ij}.
\end{equation}
We include the directed edge $i \rightarrow j$ in $\bar{G}^{(s)}_k$ whenever
$C^{(s)}_{k,ij} > 0$, and we attach $C^{(s)}_{k,ij}$ and $A^{(s)}_{k,ij}$ as
the edge count and total amount for that window.

We apply the topology feature map to each aggregated rolling-window graph:
\begin{equation}
    x^{(s)}_{k,\mathrm{topo}}
    =
    \Phi_{\mathrm{topo}}\!\left(\bar{G}^{(s)}_k\right),
    \qquad
    k=1,\ldots,K.
\end{equation}
Here, $\Phi_{\mathrm{topo}}$ denotes the full graph-to-topology pipeline: metric construction, persistent-homology computation, and persistence-diagram summarization. We collect the resulting window-level feature vectors into the topology feature sequence
\begin{equation}
    X^{(s)}_{\mathrm{topo}}
    =
    \left(
    x^{(s)}_{1,\mathrm{topo}},
    \ldots,
    x^{(s)}_{K,\mathrm{topo}}
    \right).
\end{equation}
The next three subsections define this map step by step: we construct a finite account-distance matrix from each window graph, compute $H_0$ and $H_1$ persistent homology, and summarize the resulting persistence diagrams as fixed-length numerical features.

\subsection{Distance-matrix construction}

Persistent homology requires a finite notion of distance between accounts. For each rolling-window graph, we construct this distance matrix in three steps: we combine reciprocal transaction activity, convert interaction strengths into edge lengths, and compute shortest-path distances.

We first convert the directed rolling-window graph into an undirected interaction graph. Let $S^{(s)}_{k,ij}$ denote the aggregate interaction strength between accounts $i$ and $j$ in window $k$. In this benchmark, we use transaction amount as the interaction strength and define
\begin{equation}
    S^{(s)}_{k,ij}
    =
    A^{(s)}_{k,ij}
    +
    A^{(s)}_{k,ji}.
\end{equation}
This symmetrization treats activity in either direction as evidence of interaction between the two accounts.

For each unordered account pair with $S^{(s)}_{k,ij}>0$, we convert interaction strength into an edge length:
\begin{equation}
    \ell^{(s)}_{k,ij}
    =
    \frac{1}{\log\!\left(1+S^{(s)}_{k,ij}\right)+\epsilon},
    \label{eq:length-transform}
\end{equation}
where $\epsilon>0$ prevents division by zero. This transformation maps stronger relationships to shorter edge lengths, while the logarithm reduces the influence of very large transaction amounts. Account pairs with $S^{(s)}_{k,ij}=0$ do not receive a direct edge in the undirected length graph.

Let $H^{(s)}_k$ denote the undirected length graph obtained from $\bar{G}^{(s)}_k$ after this symmetrization and length assignment. We compute pairwise account distances using shortest paths on $H^{(s)}_k$:
\begin{equation}
    D^{(s)}_{k,ij}
    =
    \operatorname{dist}_{H^{(s)}_k}(i,j).
\end{equation}
If two accounts are disconnected within the window, their shortest-path distance is infinite. We replace these infinite distances with a finite cap that is larger than the largest observed finite distance in the same window. This capped matrix preserves the interpretation that disconnected accounts are far apart while allowing the persistent-homology computation to proceed.

\subsection{Persistent-homology computation}

For each capped distance matrix $D^{(s)}_k$, we compute Vietoris--Rips persistent homology up to dimension one~\cite{edelsbrunner2002,zomorodian2005,ghrist2008,carlsson2009}. This produces persistence diagrams for $H_0$ and $H_1$. The $H_0$ diagram records how connected components merge across filtration scales, while the $H_1$ diagram records loop-like or cycle-like structure across those scales.

The $H_1$ information is especially relevant for the sparse-ring benchmark because the target pattern is cycle-based. We use these features as graph-level summaries of cycle-like geometry in each rolling-window graph. A stronger or more persistent $H_1$ signal can indicate that the window graph has acquired more loop-like structure, although this signal is computed from the symmetrized distance representation rather than from the directed ring itself.

After computing the persistence diagrams, we convert them into fixed-length numerical summaries. Section~\ref{sec:topology-feature-vector} defines the resulting topology feature vector and explains how we construct the 29 features for each rolling window.

\subsection{Topology feature vector}
\label{sec:topology-feature-vector}

For each rolling window, we summarize the topology pipeline by a fixed-length feature vector
\begin{equation}
    x^{(s)}_{k,\mathrm{topo}} \in \mathbb{R}^{29}.
\end{equation}
We construct this vector by concatenating three groups of features:
\begin{equation}
    x^{(s)}_{k,\mathrm{topo}}
    =
    \left[
    g^{(s)}_k,\,
    h^{(s)}_{k,0},\,
    h^{(s)}_{k,1}
    \right],
\end{equation}
where $g^{(s)}_k \in \mathbb{R}^{9}$ contains metric graph summaries, and
$h^{(s)}_{k,0}, h^{(s)}_{k,1} \in \mathbb{R}^{10}$ contain persistence-diagram summaries for
$H_0$ and $H_1$, respectively.

The metric graph block $g^{(s)}_k$ records nine summaries of the capped distance matrix and its associated window graph: number of nodes, number of edges, graph density, number of connected components, fraction of finite pairwise distances, mean finite distance, standard deviation of finite distances, maximum finite distance, and the finite cap used for disconnected pairs.

For each homology dimension $d \in \{0,1\}$, we compute the same ten persistence-diagram summaries: number of finite persistence features, total persistence, maximum persistence, mean persistence, standard deviation of persistence, persistence entropy, mean birth scale, standard deviation of birth scales, mean death scale, and standard deviation of death scales. The $H_0$ block summarizes connected-component structure, while the $H_1$ block summarizes loop-like or cycle-like structure.

Thus the topology-only representation contains
\begin{equation}
    9 + 10 + 10 = 29
\end{equation}
features per rolling window. For a sequence with $K$ rolling windows, the topology-only model input has dimension
\begin{equation}
    X^{(s)}_{\mathrm{topo}} \in \mathbb{R}^{K \times 29}.
\end{equation}

We use this topology sequence directly in the topology-only representation and combine it with identity-preserving graph features in the hybrid representation described next.

\section{Feature representations}

After constructing rolling-window graphs and topology summaries, we compare three feature representations. We keep the rolling windows and sequence-level labels fixed, and we vary only the information provided to the model. For each simulated sequence $s$ and representation $r$, we write the model input as
\begin{equation}
    X^{(s)}_r
    =
    \left(
    x^{(s)}_{1,r},
    \ldots,
    x^{(s)}_{K,r}
    \right),
    \qquad
    r \in \{\mathrm{raw},\mathrm{topo},\mathrm{hybrid}\}.
\end{equation}

We use three representations.

\begin{itemize}
    \item \textbf{Raw graph features.} The raw representation preserves account-pair identity. For each rolling-window graph $\bar{G}^{(s)}_k$, we flatten the directed transaction count and amount matrices and append graph-level summary statistics. In the six-account sparse benchmark, this gives
    \begin{equation}
        x^{(s)}_{k,\mathrm{raw}} \in \mathbb{R}^{103}.
    \end{equation}
    This representation captures which accounts transact, in which direction, and with what aggregate count and amount.

    \item \textbf{Topology-only features.} The topology-only representation uses the metric and persistent-homology summaries from Section~\ref{sec:topology-feature-vector}. For each rolling window,
    \begin{equation}
        x^{(s)}_{k,\mathrm{topo}} \in \mathbb{R}^{29}.
    \end{equation}
    This representation isolates global graph shape, including connectivity and loop-like structure, and lets us test how much information topology carries on its own.

    \item \textbf{Hybrid features.} The hybrid representation concatenates the raw graph features and the topology features:
    \begin{equation}
        x^{(s)}_{k,\mathrm{hybrid}}
        =
        \left[
        x^{(s)}_{k,\mathrm{raw}},
        x^{(s)}_{k,\mathrm{topo}}
        \right].
    \end{equation}
    In the six-account sparse benchmark,
    \begin{equation}
        x^{(s)}_{k,\mathrm{hybrid}} \in \mathbb{R}^{132},
    \end{equation}
    since $103+29=132$. This representation keeps identity-preserving transaction information while adding topological context.
\end{itemize}

Thus, for each simulated sequence $s$, the three model inputs have dimensions
\begin{equation}
    X^{(s)}_{\mathrm{raw}} \in \mathbb{R}^{K \times 103},
    \qquad
    X^{(s)}_{\mathrm{topo}} \in \mathbb{R}^{K \times 29},
    \qquad
    X^{(s)}_{\mathrm{hybrid}} \in \mathbb{R}^{K \times 132}.
\end{equation}
Because all three representations use the same rolling windows and the same label $y^{(s)}$, this construction lets us compare how identity-preserving graph features, topological summaries, and their combination affect sequence-level sparse-ring detection.

\section{Models}

We evaluate two sequence-level classifiers: a GRU baseline and quantum-inspired CML. Each model receives a sequence of rolling-window feature vectors and outputs one score for the full simulated sequence. For a simulated sequence $s$ and feature representation
$r \in \{\mathrm{raw},\mathrm{topo},\mathrm{hybrid}\}$, the model input is
\begin{equation}
    X^{(s)}_r \in \mathbb{R}^{K \times d_r},
\end{equation}
where $K$ is the number of rolling windows and $d_r$ is the feature dimension of representation $r$.

For model class $m$, we write the sequence-level score as
\begin{equation}
    z^{(s)}_{m,r}
    =
    f_{m,r}\!\left(X^{(s)}_r\right)
    \in \mathbb{R}.
\end{equation}
Larger values of $z^{(s)}_{m,r}$ indicate stronger model evidence that sequence $s$ contains a completed sparse ring. When we need a normalized probability-like score, we apply the logistic sigmoid,
\begin{equation}
    \sigma(u)
    =
    \frac{1}{1+\exp(-u)},
\end{equation}
and define
\begin{equation}
    \hat{p}^{(s)}_{m,r}
    =
    \sigma\!\left(z^{(s)}_{m,r}\right).
\end{equation}

The ranking metrics used in this study, including \rocauc{} and \prauc{}, depend only on the ordering of model scores. Therefore, computing these metrics from $z^{(s)}_{m,r}$ or from the monotone transformation $\hat{p}^{(s)}_{m,r}$ gives the same ranking-based evaluation. For threshold-based quantities, such as false-alarm rates, we select and apply the threshold on the same score scale.

\subsection{GRU baseline}

We use a gated recurrent unit (GRU) classifier as the standard recurrent baseline~\cite{cho2014}. The GRU reads the rolling-window feature sequence in temporal order and updates a hidden state at each step. For representation $r$, we write the recurrence as
\begin{equation}
    h^{(s)}_{k,r}
    =
    \mathrm{GRU}_r\!\left(
        x^{(s)}_{k,r},
        h^{(s)}_{k-1,r}
    \right),
    \qquad
    k=1,\ldots,K.
\end{equation}
After the GRU has processed all $K$ rolling windows, we map the final hidden state to a sequence-level score:
\begin{equation}
    z^{(s)}_{\mathrm{GRU},r}
    =
    g_{\mathrm{GRU},r}\!\left(h^{(s)}_{K,r}\right).
\end{equation}

This baseline tests whether a conventional recurrent model can use the temporal ordering of rolling-window graph features to identify completed sparse rings. We train and evaluate the GRU separately for each feature representation, so the comparison across raw, topology-only, and hybrid inputs reflects the information provided by each representation.

\subsection{Contextual Machine Learning}

We use Infleqtion's Contextual Machine Learning (CML) as the second sequence-level classifier. CML is a quantum-inspired machine learning approach for long-context sequence modeling~\cite{infleqtionCML}, motivated by links between quantum correlations, contextuality, generative modeling, and sequence learning~\cite{gao2022enhancing,PRXQuantum.4.020338,anschuetz2026arbitrary}. In this benchmark, we evaluate CML as a classical contextual classifier that maps rolling-window feature sequences to sequence-level sparse-ring scores.

The sparse-ring task provides a natural test case for contextual classification. No single transaction edge determines the label. Instead, the model must relate directed edges that appear in different rolling windows and determine whether they complete a coordinated ring. The relevant evidence is therefore distributed across time and graph structure, and the significance of one edge depends on the surrounding transaction history.

For representation $r$, CML receives the same model input as the GRU,
\begin{equation}
    X^{(s)}_r \in \mathbb{R}^{K \times d_r},
\end{equation}
and returns a sequence-level score
\begin{equation}
    z^{(s)}_{\mathrm{CML},r}
    =
    f_{\mathrm{CML},r}\!\left(X^{(s)}_r\right).
\end{equation}
In this study, we evaluate the classical quantum-inspired CML implementation as a contextual sequence model. Quantum-native hardware extensions are left as a future direction.

\subsection{Controlled model comparison}
We train and evaluate the GRU and CML models under matched conditions. For each dataset regime and feature representation
$r \in \{\mathrm{raw},\mathrm{topo},\mathrm{hybrid}\}$, both models use the same training, validation, and test splits, the same rolling-window input sequences $X^{(s)}_r$, and the same sequence-level labels $y^{(s)}$. Thus, holding $r$ fixed, the comparison isolates the effect of model class.

Because GRU and CML scores may have different numerical scales, we compare the models through held-out evaluation metrics rather than through raw score magnitudes. For each representation $r$, we compare
\begin{equation}
    A_{\mathrm{PR}}(\mathrm{GRU},r)
    \quad \text{and} \quad
    A_{\mathrm{PR}}(\mathrm{CML},r),
\end{equation}
together with the corresponding ROC-AUC values and decoy false-alarm rates. This matched design tests whether quantum-inspired CML can use the same temporal, graph, and topological information as effectively as, or more effectively than, a standard recurrent baseline.

\section{Experimental design}

We evaluate the GRU and CML models on the sparse-ring benchmark regimes described above. For each regime, we generate $N=260$ simulated transaction-history sequences and compute raw, topology-only, and hybrid representations for the same underlying sequences. For representation
$r \in \{\mathrm{raw},\mathrm{topo},\mathrm{hybrid}\}$, the model input for sequence $s$ is
\begin{equation}
    X^{(s)}_r \in \mathbb{R}^{K \times d_r}.
\end{equation}
Equivalently, we view the full dataset for representation $r$ as a tensor
\begin{equation}
    \mathbf{X}_r \in \mathbb{R}^{N \times K \times d_r},
\end{equation}
where $N=260$ is the number of simulated sequences, $K=28$ is the number of rolling windows per sequence, and $d_r$ is the feature dimension. The feature dimensions are
\[
    d_{\mathrm{raw}}=103,
    \qquad
    d_{\mathrm{topo}}=29,
    \qquad
    d_{\mathrm{hybrid}}=132.
\]

We use a simulated horizon of $T=30$ days, a rolling-window length of $L=3$ days, and a step size of one day. This gives
\begin{equation}
    K = T-L+1 = 30-3+1 = 28
\end{equation}
rolling windows per sequence. In the sparse-ring benchmark, we schedule the ring edges with gaps between four and nine days. Because these gaps exceed the three-day rolling-window length, no single rolling-window graph contains the full completed ring. The model must therefore integrate evidence across multiple windows rather than detect the ring inside one static graph. Table~\ref{tab:experiment-design} summarizes the main experimental settings.

\begin{table}[tbp]
\centering
\caption{Main sparse-ring benchmark design.}
\label{tab:experiment-design}
\begin{tabular}{ll}
\toprule
Component & Setting \\
\midrule
Benchmark regimes & Easy-clean, hard-decoy, mixed \\
Simulated sequences per regime & 260 \\
Accounts per sequence & 6 \\
Simulated horizon & 30 days \\
Rolling window & 3 days, step size 1 day \\
Windows per sequence & 28 \\
Ring size & 3 accounts \\
Sparse edge gaps & 4--9 days between ring edges \\
Feature dimensions & Raw: 103, topology: 29, hybrid: 132 \\
Background base volume & 35 expected transactions/day before multipliers \\
Train/validation/test split & 70\% / 15\% / 15\%, stratified by sequence label \\
Training epochs & 80 \\
\bottomrule
\end{tabular}
\end{table}

We split each regime into training, validation, and test sets. We fit model parameters on the training set, use the validation set for model selection and threshold selection, and reserve the test set for final evaluation. We stratify the split by the sequence-level label $y^{(s)}$ so that completed-ring and non-completed-ring sequences appear in each split. For a given regime, we use the same split across feature representations and model classes, so the comparisons isolate differences due to representation and model choice.

We use test PR-AUC as the primary evaluation metric. Precision-recall evaluation is appropriate for fraud-screening benchmarks because the model produces a ranked list of risky sequences, and in practice only the highest-risk alerts may be reviewed. PR-AUC summarizes how well the model ranks completed-ring sequences above non-completed-ring sequences across decision thresholds, and it is especially useful when false alarms matter or when fixed-threshold accuracy is unstable in small exploratory datasets~\cite{saito2015precision,davis2006prroc}. We also report ROC-AUC and decoy false-alarm rates where they are needed to evaluate the research propositions.

\section{Results}

\subsection{Sparse-ring benchmark regimes}

\begin{figure}[tbp]
    \centering
    \includegraphics[width=\linewidth]{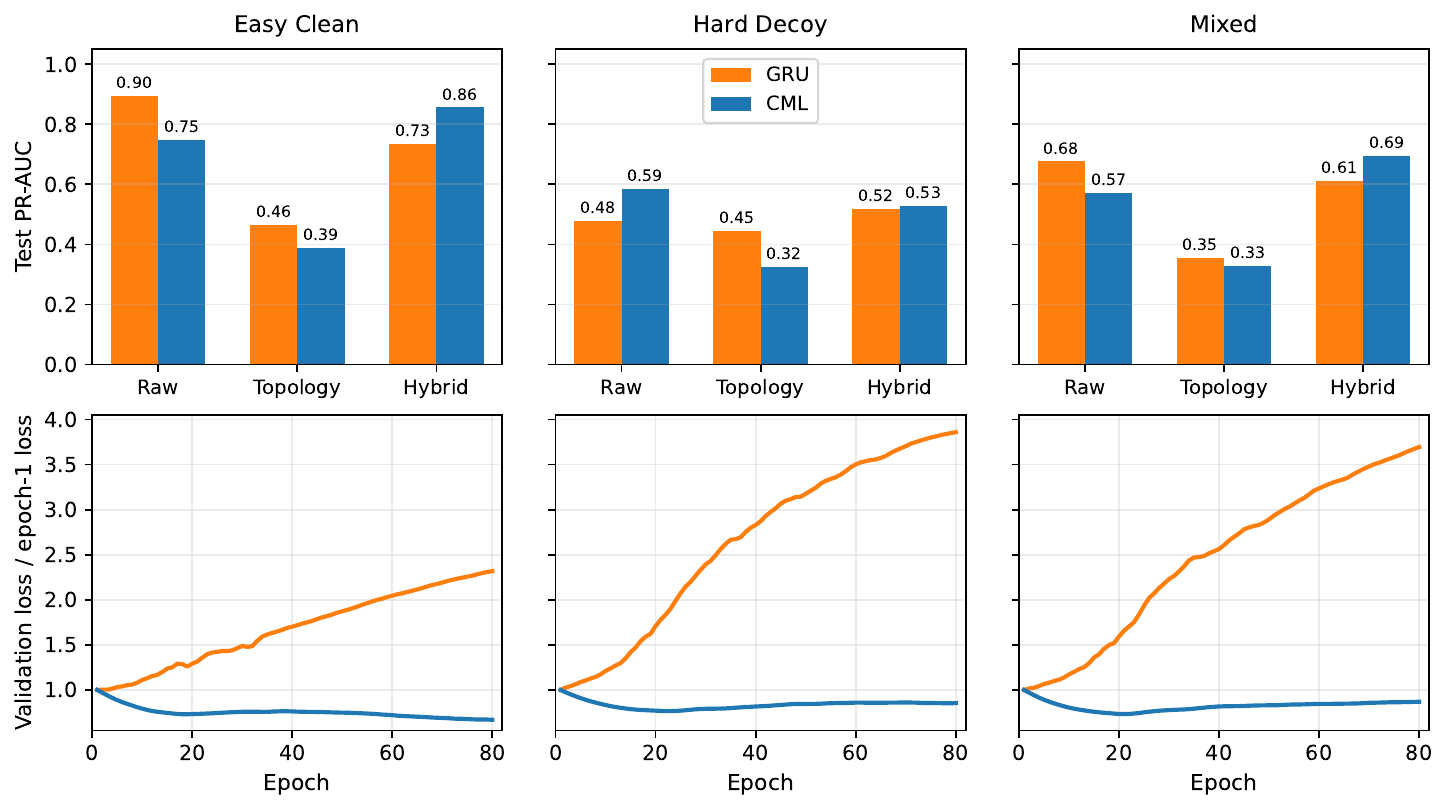}
    \caption{Sparse-ring benchmark results across the three benchmark regimes. Each column corresponds to one regime: easy-clean, hard-decoy, or mixed. The top row reports test \prauc{} for GRU and Infleqtion CML using raw graph, topology-only, and hybrid feature representations. The bottom row shows validation-loss curves for the hybrid representation, normalized by each model's epoch-1 validation loss. We use these learning curves as diagnostics; final model comparison is based on held-out ranking metrics such as test \prauc{}.}
    \label{fig:cml-diagnostics}
\end{figure}

Figure~\ref{fig:cml-diagnostics} reports the main sparse-ring benchmark results. The top row shows test \prauc{} for GRU and CML across the easy-clean, hard-decoy, and mixed regimes. The bottom row shows normalized validation-loss curves for the hybrid representation. All reported \prauc{} values are test-set values, so larger values indicate better ranking of completed-ring sequences above sequences without completed rings.

We observe three main patterns. First, broken-ring decoys make the benchmark substantially harder. In the easy-clean regime, the non-completed-ring sequences contain only normal background activity, and the best test \prauc{} values are 0.895 for the GRU with raw graph features and 0.856 for CML with hybrid features. In the hard-decoy regime, the non-completed-ring sequences contain incomplete ring-like activity, and performance drops for both models and most feature sets. This drop indicates that the hard-decoy regime tests ring-completion detection rather than simple detection of injected sparse activity.

Second, raw and hybrid representations carry more useful signal than topology-only representations for this supervised sparse-ring task. Across all three regimes, topology-only inputs produce lower \prauc{} values than raw or hybrid inputs. This result is consistent with the construction of the topology representation: it summarizes global graph shape, but it removes explicit account-pair identity and edge direction. Since the label depends on whether specific directed edges close a ring, this loss of identity information matters.

Third, the hybrid representation gives the strongest CML results in the easy-clean and mixed regimes. In the easy-clean regime, CML improves from 0.748 with raw graph features to 0.856 with hybrid features. In the mixed regime, CML improves from 0.571 with raw graph features to 0.695 with hybrid features. In the hard-decoy regime, CML performs best with raw graph features, but the hybrid CML result remains close to the hybrid GRU result. These results suggest that topology is most useful as additional structural context rather than as a replacement for identity-preserving graph features.

The bottom row of Figure~\ref{fig:cml-diagnostics} provides a training diagnostic for the hybrid representation. The GRU validation loss increases during training, especially in the hard-decoy and mixed regimes, while the CML curve remains flatter. Because the two models use different training objectives, we use these curves only as diagnostics and compare final model performance using held-out ranking metrics such as test \prauc{}.

\subsection{Feature-representation comparison}

We compare the three input feature sets to examine how much information each representation provides for sparse-ring detection. The prediction task remains fixed: the model must decide whether a sequence contains a completed sparse ring. We vary only the representation given to the model: raw graph features, topology-only features, or hybrid features.

This comparison isolates the role of feature construction. Strong topology-only performance would suggest that global topological summaries contain enough information for the sparse-ring task on their own. Strong hybrid performance would suggest that topology adds useful structural context when combined with identity-preserving graph features.

\begin{figure}[tbp]
    \centering
    \includegraphics[width=0.7\linewidth]{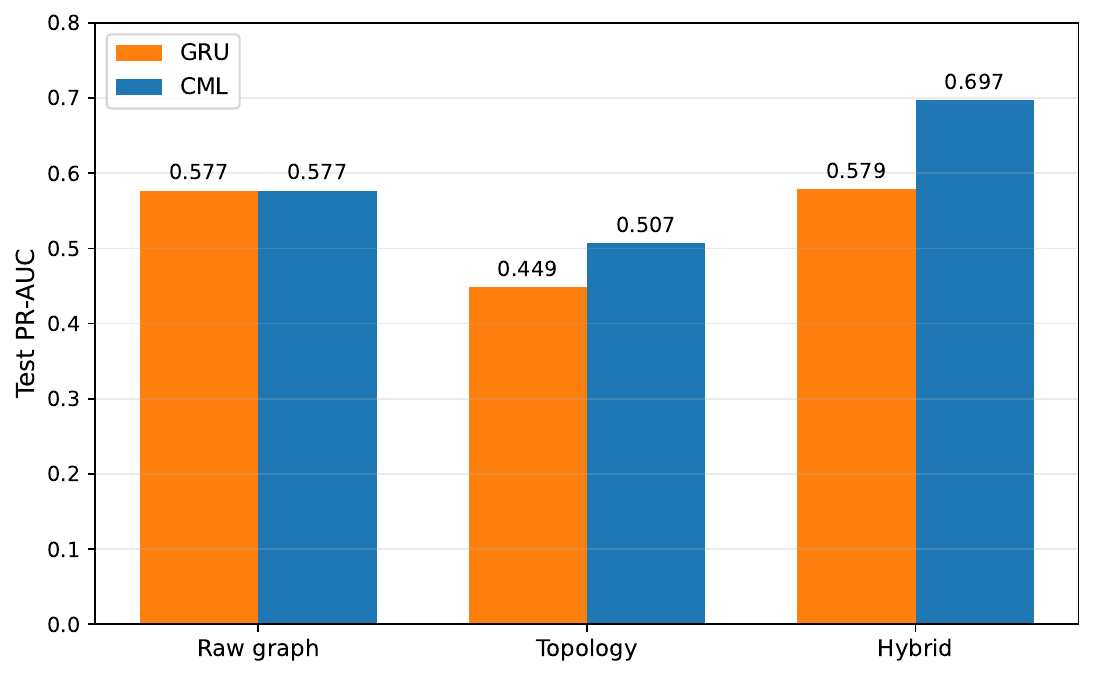}
    \caption{Feature-representation comparison using test \prauc{}. The prediction task is fixed, while the model input changes across raw graph features, topology-only features, and hybrid features. Raw graph features preserve account-pair transaction information, topology-only features use persistent-homology and metric summaries, and hybrid features combine both. The hybrid representation gives the strongest CML result, suggesting that topological summaries are most useful when they augment identity-preserving graph features.}
    \label{fig:representation-comparison}
\end{figure}

Figure~\ref{fig:representation-comparison} shows the test \prauc{} results. The hybrid representation gives the strongest CML result. Raw graph features preserve account-pair transaction information, while topology-only features summarize global graph shape through metric and persistent-homology summaries. The hybrid representation combines these two sources of information, supporting the interpretation that topology is most useful as an additional source of graph context rather than as a standalone replacement for raw transaction features.

\section{Discussion and validity considerations}

\subsection{Interpretation of the results}

The experiments point to a focused interpretation: topological summaries are most useful as contextual features when paired with identity-preserving transaction information. Persistent-homology features capture broad connectivity and loop-like structure in rolling-window graphs, while raw graph features preserve account-pair identity and edge direction. Because the sparse-ring label depends on both graph shape and specific directed transfers, the hybrid representation provides the most appropriate setting for this benchmark.

The decoy regimes clarify what the models learn. Easy-clean comparisons test whether a model can detect injected sparse activity against normal background traffic. Hard-decoy comparisons test a stricter condition: whether the model can distinguish a completed directed ring from a similar incomplete ring-like pattern. The performance drop in the hard-decoy regime shows that ring-completion detection is substantially harder than simple anomaly detection.

The topology-only results show the limits of global persistent-homology summaries for this supervised task. These features preserve information about connected components, metric structure, and loop-like persistence, but they compress away account identity and directionality. That compression is costly when the label depends on whether particular directed transfers close a particular cycle. The hybrid results therefore provide the main positive signal for topology: topological summaries can add structural context when raw graph features retain the local transaction identities needed to recognize the ring.

The model comparison gives a cautious but useful conclusion about CML. CML is competitive with the GRU baseline and obtains its strongest results in hybrid settings, especially in the easy-clean and mixed regimes. In the hard-decoy regime, hybrid CML and hybrid GRU remain close, suggesting that the current topology features help as auxiliary context but do not fully resolve the most difficult ring-completion cases. Overall, the results support continued investigation of CML as a contextual sequence model and of topology as a complementary graph feature layer.

\subsection{Benchmark scope}

We use a controlled synthetic simulator to isolate the sparse-ring detection problem. This gives us known event timing, participants, and labels for completed sparse rings and broken-ring decoys, which makes the benchmark useful for comparing feature representations and model classes under matched conditions. The next validation step is to calibrate the simulator against operational or carefully anonymized institutional data that include reporting delays, seasonal effects, heterogeneous customer behavior, changing adversarial tactics, regulatory constraints, and investigator-dependent labels.

The current benchmark uses small graphs and sequence-level labels. The small scale lets us inspect examples, compute persistent homology efficiently, and isolate whether a sparse ring has been completed. Scaling the method to larger transaction networks will require efficient feature extraction, careful subgraph or candidate-window construction, and mechanisms that preserve explanatory links to accounts, edges, and time windows. Future versions should also move toward account-level or edge-level scoring so that high-risk sequences can be localized to specific entities or transactions.

\subsection{Representation and evaluation scope}

The topology pipeline uses global persistent-homology summaries computed from symmetrized rolling-window distance matrices. These summaries capture connectivity, metric structure, and loop-like persistence, but they remove account-pair identity and edge direction. The weak topology-only results therefore identify a limitation of this particular global representation for a label defined by specific directed transfers. More local topology features, computed on ego-networks, candidate rings, or suspicious subgraphs, may retain more explanatory information.

The statistical evidence remains exploratory. The benchmark size is modest, the results are not yet averaged across repeated random seeds, and the model comparison uses representative GRU and CML implementations rather than an exhaustive hyperparameter search. We use PR-AUC as the primary ranking metric, but operational fraud screening would also require precision at top-$k$, recall at fixed alert budgets, calibration quality, investigation cost, and false-positive analysis on realistic non-fraud activity. These extensions would make the evaluation more reliable and more closely aligned with practical fraud-screening workflows.

\section{Future research}

Future work can extend this prototype while preserving the controlled structure of the sparse-ring benchmark. The main priorities are to strengthen the statistical evidence, increase the temporal difficulty of the benchmark, improve the topological representation, explore quantum-native extensions, and move toward operationally meaningful evaluation.

\subsection{Statistical reliability and long-memory benchmarks}

The immediate priority is to repeat the experiments across multiple random seeds. Future studies should report means, standard deviations, confidence intervals, and paired comparisons for \prauc{} and \rocauc{}. This would help distinguish stable feature and model effects from favorable train-test splits or particular simulation draws.

The benchmark should also test longer temporal separations between injected ring edges. The current experiments separate sparse-ring edges by gaps of several days within a relatively short horizon. A more demanding long-memory benchmark would increase the distance between ring edges and test whether each model can preserve weak evidence over longer temporal spans. This extension is especially relevant for CML, because contextual sequence models may become more useful when the discriminative signal depends on relationships between events that are far apart in time.

\subsection{Topological and higher-order graph representations}

A second direction is to improve the topological representation. The current pipeline computes persistent homology on the full rolling-window graph. Future experiments should also compute topological summaries on ego-networks, candidate rings, suspicious subgraphs, or neighborhoods selected by a preliminary graph model. These more local constructions could retain more entity-level information and make topological features easier to connect to investigator-facing explanations.

Future work should also evaluate richer persistence-diagram representations, including persistence images, persistence landscapes, silhouettes, Betti curves, and learned topological embeddings~\cite{adams2017persistenceimages,bubenik2015landscapes,chazal2014silhouettes}. These representations may preserve more information than the hand-crafted persistence summaries used in the current prototype.

\subsection{Quantum-native extensions}

A further direction is to extend the quantum component of the study. The present experiments use Infleqtion CML as a classical, quantum-inspired model. Future research should investigate settings in which quantum hardware or quantum-native subroutines could contribute directly. Candidate directions include quantum-enhanced contextual learning~\cite{gao2022enhancing,PRXQuantum.4.020338,anschuetz2026arbitrary} and quantum topological data analysis (QTDA)~\cite{lloyd2016quantum,mcardle2026streamlined}.

Recent work on quantum--topological signal processing (QTSP)~\cite{leditto2025tsp,leditto2025quantum} and higher-order network analysis also suggests a route for modeling transaction activity as signals on edges or simplicial complexes. In that setting, Hodge-based filtering could emphasize cyclic or coordinated flow structures. This direction is closely aligned with sparse-ring fraud, where the relevant signal is not only node activity but also whether directed transfers form a coordinated cycle.

\subsection{Decoys, real data, and operational evaluation}

The decoy-generation process should become more diverse. Broken rings are a useful first hard-negative case, but future benchmarks should also include partial cycles, reciprocal flows, high-value bursts, repeated near-miss structures, benign merchant-like cycles, and adversarial patterns that intentionally mimic ordinary activity. Stronger decoys would reduce the risk that models learn superficial simulation artifacts instead of the intended temporal mechanism.

Future work should also evaluate the method on real, anonymized, or institutionally calibrated transaction data and move toward entity-level scoring. After a risky sequence or window is detected, the system should rank the accounts and edges that contribute most to the signal. Future evaluations should therefore report alert volume, precision at top-$k$, recall at fixed review capacity, calibration quality, and the stability of explanations across nearby windows. These metrics would connect the prototype more directly to how a fraud-screening system would be assessed in practice.

\section{Conclusion}

We study dynamic fraud detection as a temporal graph problem. We introduce a controlled sparse-ring benchmark in which coordinated fraud appears as a completed directed cycle whose edges are distributed across multiple days. This benchmark lets us test whether sequence-level models can distinguish completed rings from ordinary background activity and from broken-ring decoys that contain similar sparse transfers but do not close the cycle.

The results refine the role of topology in this setting. Persistent-homology summaries alone are too compressed for the supervised sparse-ring task, because they remove account-pair identity and edge direction. However, they can provide useful structural context when combined with identity-preserving transaction features. The strongest evidence therefore supports a hybrid representation that combines raw graph features with persistent-homology summaries, especially under decoy-based benchmark conditions.

The model comparison also gives a focused conclusion about quantum-inspired CML. CML is competitive with the GRU baseline and achieves its strongest results in hybrid feature settings, suggesting that contextual sequence models are worth further study when fraud evidence is distributed across time and graph structure. More broadly, the sparse-ring benchmark provides a useful testbed for studying temporal graph structure, topological augmentation, and contextual learning. Future work can strengthen this prototype through repeated-seed evaluation, longer temporal gaps, local topology, richer decoy families, entity-level explanation, and quantum-native extensions such as quantum contextual learning, QTDA, and quantum--topological signal processing.

\bibliographystyle{unsrt}
\bibliography{ref}

\end{document}